%% file: neurips_2026.tex
\documentclass{article}

\usepackage[preprint]{neurips_2026}
\usepackage{multirow}
\usepackage{amsmath, amssymb, amsthm}
\usepackage{graphicx}
\usepackage{makecell}
\input{math_commands.tex}

\usepackage[utf8]{inputenc} 
\usepackage[T1]{fontenc}    
\usepackage{hyperref}       
\usepackage{url}            
\usepackage{booktabs}       
\usepackage{amsfonts}       
\usepackage{nicefrac}       
\usepackage{microtype}      
\usepackage{xcolor}         

\title{Beyond Rigid Geometries: The Spline-Pullback Metric for Universal Diffeomorphic SPD Representation Learning}

\author{Tushar Das \\
National Institute of Technology Jamshedpur \\
\texttt{tdas2663@gmail.com} \\
\And
 Subrata Dutta \\
National Institute of Technology Jamshedpur \\
\texttt{sdutta.cse@nitjsr.ac.in}
\And
Sarmistha Neogy \\
Jadavpur University \\
\texttt{sarmistha.neogy@jadavpuruniversity.in}
\And
Koushlendra Kumar Singh \\
Machine Vision and Intelligence Lab \\
National Institute of Technology Jamshedpur\\
\texttt{koushlendra.cse@nitjsr.ac.in}}

\begin{document}

\maketitle

\begin{abstract}
The integration of Symmetric Positive Definite (SPD) matrices into deep learning has historically relied on fixed algebraic Riemannian metrics. Analogous to hand-crafted features in classical machine learning, these static formulations impose rigid geometries limiting network expressivity and adaptability. Recent attempts to parameterize these geometries often violate the axioms of primary matrix functions through unconstrained powers or rank-dependent scaling, inviting spatial folding, loss of global surjectivity, and gradient collapse at spectral singularities. In this paper, we introduce the Spline-Pullback Metric (SPM), instantiated as Spectral-SPM and Cholesky-SPM, marking a paradigm shift from static metric selection to universal geometric approximation. By parameterizing the global diffeomorphism via a rank-invariant, monotonically constrained B-spline, SPM acts as a dense universal approximator for strictly increasing $C^1$ diffeomorphisms and theoretically subsumes existing pullback metrics while enabling localized non-linear spectral modelling. Topologically, SPM provides a globally bijective pullback geometry precluding rank-swapping discontinuities and gradient instabilities. Empirically, SPM achieves a state-of-the-art performance across 3 datasets utilizing Linear Probes, SPDNets, and deep Riemannian ResNets.
\end{abstract}
\section{Introduction}

Covariance matrices, represented as Symmetric Positive Definite (SPD) matrices, constitute a key component in numerous modern deep learning applications spanning areas such as radar signal processing \citep{brooks2019riemannian} and electroencephalography \citep{li2025spdim}, computer vision \citep{huang2017riemannian}, and medical imaging \citep{chakraborty2020manifoldnet}. From the geometrical point of view, the SPD matrix space $\mathcal{S}_{++}^n$ represents an open and convex cone within the ambient space of symmetric matrices $\mathcal{S}^n$. As a purely topological space, this cone possesses no inherent geometry or spatial curvature. Because applying standard Euclidean operations directly to SPD matrices violates the positivity boundary constraint, the manifold must be endowed with a Riemannian metric to induce a valid geometry, thereby defining geodesic distances and enabling mathematical projections

The problem of handling the positivity boundary constraint has been traditionally tackled by formulating appropriate geometry manually. Among the pioneering methods, the Affine-Invariant Riemannian Metric (AIRM) \citep{pennec2006riemannian} and the Log-Euclidean Metric (LEM) \citep{arsigny2006log} were followed by the later developed Power-Euclidean \citep{dryden2009non}, Log-Cholesky (LCM) \citep{lin2019riemannian}, and Bures-Wasserstein metrics \citep{bhatia2019bures} and various generalizations of this metric \citep{han2023learning}. Endowing the manifold with such metrics has paved the way for extending key deep learning building blocks to the SPD domain. Examples of such include Riemannian Multinomial Logistic Regression (MLR) \citep{chen2024riemannian}, convolutions, Riemannian batch normalization \citep{brooks2019riemannian,chen2024gyrogroup}, residual blocks \citep{katsman2024riemannian}, and graph neural networks\citep{cruceru2021computationally}. However, relying on classical metric selection remains a manual process, akin to handcrafted feature selection in traditional Machine Learning. The central challenge in advancing SPD matrix and Riemannian metric learning is transitioning from these static mathematical priors to dynamic, computationally efficient, and completely learnable topological representations.

\textbf{Literature Limitations:}
Although implicit pullback metrics like LEM and LCM preclude the expensive matrix inversions and gradient instability of AIRM, they enforce rigid global topological deformations indifferent to localized spectral noise. Recent advancements attempt to parameterize Riemannian metrics to improve expressivity, but often rely on mathematically ill-posed transformations. For instance, rank-dependent mappings like ALEM \citep{chen2024adaptive} violate the Primary Matrix Functions definition \citep{bhatia1997matrix} due to eigenvector non-uniqueness, causing one-to-many topological failures. Conversely, polynomial power mappings like PCM \citep{chen2026fast} lack global surjectivity onto unconstrained Euclidean space. To prevent undefined negative pre-images, PCM relies on non-injective computational clipping (e.g., $\max(x, \epsilon)$), breaking the bijectivity strictly required for a geodesically complete Riemannian metric. We provide a rigorous mathematical proof of these diffeomorphic failures in the Appendix.

Furthermore, adapting Euclidean diffeomorphism approximators (like Affine Coupling Flows \citep{teshima2020coupling} or deep MLPs \citep{hwang2026minimum}) is infeasible for Riemannian metric learning. They typically require $C^2$ smooth targets and only guarantee $C^0$ (uniform) convergence, leaving the first derivatives, essential for stable Riemannian back-propagation, structurally discontinuous, often requiring computationally heavy layer cascades.

To resolve all these limitations, we introduce the \textbf{Spline-Pullback Metric (SPM)} and its two instantiations, Spectral-SPM (S-SPM) and Cholesky-SPM (C-SPM). In contrast to ALEM or PCM, we mathematically prove that SPM metric itself is a fully learnable, geodesically complete global Riemannian diffeomorphism and a universal approximator. Requiring only a $C^1$ target, SPM guarantees the $C^1$ convergence necessary for stable back-propagation in a single, parameter-efficient layer. To the best of our knowledge, SPM is the first framework to introduce the paradigm of universal Riemannian diffeomorphic metric
learning parameterized via B-splines, capable of localized spectral modelling without sacrificing the fundamental axioms of differential geometry.

Our main contributions are summarized as follows:
\begin{itemize}
    \item We mathematically formalize and prove the desirable properties of SPM and its instantiations S-SPM and C-SPM.
    \item We highlight the current metrics' topological limitations and SPM's universal solution.
    \item We introduce an asymmetric Float64 matrix perturbation protocol to bound the Lipschitz constant of the Daleck\u{\i}\u{\i}-Kre\u{\i}n structural gradient, solving the numerical instability of Riemannian backpropagation.
    \item Under a rigorous evaluation protocol, SPM achieves state-of-the-art accuracy across all 9 tested combinations of datasets and architectures, across 7 metrics.
\end{itemize}

\section{Preliminaries}
\label{sec:preliminaries}

We establish the notation and geometric concepts for Symmetric Positive Definite (SPD) matrices and the mechanism of Riemannian isometric pullbacks, which form the theoretical underpining of our framework.

We denote the space of $n \times n$ real symmetric matrices as $\mathcal{S}^n$, which forms a flat Euclidean vector space equipped with the standard Frobenius inner product, $\langle X, Y \rangle_F = \text{tr}(X^\top Y)$. 

The set of $n \times n$ Symmetric Positive Definite (SPD) matrices, denoted as $\mathcal{S}_{++}^n$, is an open, convex cone within $\mathcal{S}^n$. It is worth noting that the topological space $\mathcal{S}_{++}^n$ possesses no inherent spatial curvature of its own; rather, its geometry is induced by the choice of the Riemannian metric tensor $g$ placed upon it. Because $\mathcal{S}_{++}^n$ is an open cone rather than a complete vector space, applying standard Euclidean linear operations directly to SPD matrices may produce matrices that violate the strict positivity constraint (falling outside the manifold) or results in the well-documented "swelling effect" \citep{arsigny2006log}. Consequently, $\mathcal{S}_{++}^n$ must be endowed with a Riemannian metric to measure geodesic distances and define valid tangent space projections. At any given point $P \in \mathcal{S}_{++}^n$, its tangent space $\mathcal{T}_P\mathcal{S}_{++}^n$ is isomorphic to the flat space of symmetric matrices $\mathcal{S}^n$.

To facilitate operations requiring hierarchical sensor ordering, we also define the Cholesky space. Let $\mathcal{L}^n$ denote the flat Euclidean space of $n \times n$ lower triangular matrices. The Cholesky manifold, denoted as $\mathcal{L}_{++}^n$, is defined as the open submanifold of $\mathcal{L}^n$. It consists of lower triangular matrices where all diagonal elements are strictly positive \citep{lin2019riemannian}. Any SPD matrix $S \in \mathcal{S}_{++}^n$ can be uniquely mapped to the Cholesky manifold via the Cholesky decomposition, $S = L L^\top$, where $L \in \mathcal{L}_{++}^n$.

While historically popular classical metrics like the \textbf{Affine-Invariant Riemannian Metric (AIRM)} \citep{pennec2006riemannian} induce negative sectional curvature requiring expensive iterative approximations \citep{karcher1977riemannian}, modern networks have shifted toward pullback metrics that induce a globally flat geometry.

\textbf{Definition 1 (Pullback Euclidean Metric)\citep{thanwerdas2022geometry}.} Let $\mathcal{M}$ represent a smooth topological manifold (such as the SPD cone $\mathcal{S}_{++}^n$), and let $\mathcal{E}$ represent a flat Euclidean vector space (such as $\mathcal{S}^n$ or $\mathcal{L}^n$). If there exists a smooth, globally bijective mapping, a diffeomorphism, $\phi: \mathcal{M} \to \mathcal{E}$, one can equip $\mathcal{M}$ with a globally flat Riemannian geometry by pulling back the standard Frobenius inner product from $\mathcal{E}$. The Pullback Euclidean Metric (PEM) $g_\phi$ evaluated at point $P \in \mathcal{M}$ for tangent vectors $V_1, V_2 \in \mathcal{T}_P\mathcal{M}$ is defined point-wise as:
\begin{equation}
    g_\phi|_P(V_1, V_2) = \langle \mathcal{D}\phi_P(V_1), \mathcal{D}\phi_P(V_2) \rangle_F
\end{equation}
where $\mathcal{D}\phi_P : \mathcal{T}_P\mathcal{M} \to \mathcal{E}$ is the pushforward differential map of $\phi$ at $P$.

Because the target space $\mathcal{E}$ is flat, any Riemannian metric constructed via this global diffeomorphic pullback guarantees that the manifold $\mathcal{M}$ possesses zero sectional curvature globally. This eliminates the need for iterative Karcher flow approximations.

The topological deformation of a Pullback Euclidean Metric is dictated by $\phi$. Two prevalent baselines enforce immutable logarithmic deformations:

\textbf{Log-Euclidean Metric (LEM) \citep{arsigny2006log}:} Maps to $\mathcal{S}^n$ via the eigendecomposition $S = U \Sigma U^\top$:
\begin{equation}
    \phi_{LE}(S) = \log(S) = U \ln(\Sigma) U^\top
\end{equation}

\textbf{Log-Cholesky Metric (LCM) \citep{lin2019riemannian}:} Maps to the Cholesky manifold $\mathcal{L}_{++}^n$ via factorization $S = L L^\top$ to bypass $\mathcal{O}(n^3)$ eigendecompositions. It applies the natural logarithm exclusively to the strictly positive diagonal elements:
\begin{equation}
    \phi_{LC}(S) = \lfloor L \rfloor + \log(\text{diag}(L))
\end{equation}
where $\lfloor L \rfloor$ isolates the strictly lower triangular elements. 

While stable, LEM and LCM enforce rigid deformations. A discussion on fundamental Riemannian operators and Splines is deferred to the Appendix.

\section{The Spline-Pullback Metric framework (SPM)}
\label{sec:method}

Classical metrics impose topological restrictions, and scalar parametric metrics often falter at boundaries. We propose the Spline-Pullback Metric (SPM) framework to address these issues and propose a universally applicable framework. SPM learns non-linear topological deformations end-to-end by drawing on the localized support properties of B-splines. In this section, we formalize the universal scalar diffeomorphism at the core of SPM, detail its matrix instantiations, and outline the resulting geometric properties. The Curry-Schoenberg Theorem \citep[Ch.~IX, Thm.~44]{deboor2001} establishes that B-splines provide an optimal, linearly independent basis for piecewise polynomial spaces, enabling non-linear modelling without triggering the Runge's phenomenon that mars polynomial regression. The proofs of all the subsequent theorems are provided in the Appendix.

\subsection{The universal scalar diffeomorphism}

 SPM relies on a parameterized, strictly monotonic scalar mapping $f_\theta : \mathbb{R}_{>0} \to \mathbb{R}$. To avert the lack of global surjectivity and gradient instability that plague power metrics, and ensure full domain coverage, $f_\theta(x)$ operates in the logarithmic domain and combines a central B-spline with bounded linear extrapolations at the edges.

\textbf{Definition 2 (The SPM Scalar Generator).}
\textit{Let $T = \{t_0, t_1, \dots, t_m\}$ be a predefined, clamped knot vector. The SPM scalar generator $f_\theta : \mathbb{R}_{>0} \to \mathbb{R}$ is defined as a composite mapping, evaluated as a linear combination of B-spline basis functions $B_{i,k}(x)$ of degree $k$. Outside $T$, it extends via bounded linear extrapolation using the boundary derivatives.}
\begin{equation}
f_\theta(x) = 
\begin{cases} 
m_{L}(\log(x) - t_k) + S(t_k) & \text{if } \log(x) < t_k \\
S(\log(x)) & \text{if } t_k \le \log(x) \le t_{m-k} \\
m_{R}(\log(x) - t_{m-k}) + S(t_{m-k}) & \text{if } \log(x) > t_{m-k} 
\end{cases}
\end{equation}
\textit{where $S(x) = \sum c_i B_{i,k}(x)$ is a B-spline of degree $k$ with learnable control coefficients $c_i \in \mathbb{R}$, and $m_L, m_R$ represent the strictly positive analytical derivatives of $S(x)$ at the respective grid boundaries.}

The basis functions $B_{i,k}(x)$ are defined via the numerically stable Cox-de Boor recursion formula\citep[Ch.~IX, Prop.~i]{deboor2001}. 
For degree $k=0$:
\begin{equation}
B_{i,0}(x) = 
\begin{cases} 
1 & \text{if } t_i \le x < t_{i+1} \\
0 & \text{otherwise}
\end{cases}
\end{equation}
And for degree $k > 0$, they are constructed recursively:
\begin{equation}
B_{i,k}(x) = \frac{x - t_i}{t_{i+k} - t_i} B_{i,k-1}(x) + \frac{t_{i+k+1} - x}{t_{i+k+1} - t_{i+1}} B_{i+1,k-1}(x)
\end{equation}

\textbf{Proposition 1 (B-Spline Derivative).} \textit{As a direct consequence of the Cox-de Boor recursion, the first derivative of the B-spline $S(x)$ of degree $k$ is another B-spline of degree $k-1$: \citep[Ch.~X, Prop.~viii]{deboor2001}}
\begin{equation}
S'(x) = \sum_{i} k \frac{c_i - c_{i-1}}{t_{i+k} - t_i} B_{i,k-1}(x)
\end{equation}

\textbf{Theorem 1 (Global Scalar Diffeomorphism).} \label{thm:global scalar diff}
\textit{\ref{sec:global scalar diff proof} If the sequence of learnable coefficients $c$ is strictly monotonically increasing ($c_i > c_{i-1}$ for all $i$), then $f_\theta(x)$ is a globally bijective, $C^1$-smooth diffeomorphism from $\mathbb{R}_{>0}$ to $\mathbb{R}$.}

To apply this constraint directly during continuous gradient descent, we use a cumulative \textit{softplus} reparameterization: $c_i = c_{i-1} + \text{softplus}(w_i) + \epsilon$. While classical metrics enforce rigid deformations, a fully learnable metric must possess the theoretical capacity to model any valid non-linear metric.

\textbf{Theorem 2 (Universal Approximation on Compact Spectra).} \label{thm:univ approx}
\textit{\ref{sec:univ approx proof} The constrained SPM mapping $f_\theta(x)$ is dense in the space of strictly increasing $C^1$ diffeomorphisms from the positive real line to the unconstrained real line, denoted $\text{Diff}_+^1(\mathbb{R}_{>0}, \mathbb{R})$, over any compact spectral interval. That is, for any target geometry $g \in \text{Diff}_+^1(\mathbb{R}_{>0}, \mathbb{R})$, any compact data interval $[a,b] \subset \mathbb{R}_{>0}$, and any error tolerance $\epsilon > 0$, there exists a finite knot vector $T$ encompassing $\log([a,b])$ and a strictly increasing sequence of control coefficients $c_i > c_{i-1}$ such that both $\sup_{x \in [a,b]} |f_\theta(x) - g(x)| < \epsilon$ and $\sup_{x \in [a,b]} |f_\theta'(x) - g'(x)| < \epsilon$.}

A direct mathematical consequence of this dense approximation capacity is that SPM unifies prior metric literature into a single framework.

\textbf{Corollary 1 (Universal Subsumption of Analytical Geometries).}\label{cor:univ subsump}
\textit{\ref{sec:univ subsump proof}
Let $\mathcal{H}_{SPM}$ denote the hypothesis space of Riemannian metrics induced by the SPM framework. By virtue of Theorem 2, $\mathcal{H}_{SPM}$ theoretically subsumes all orientation-preserving, topologically valid parametric and non-parametric pullback metrics currently established in the literature.}

\textbf{Proposition 2 (Topological Safety and Geometric Expressivity of the SPM Generator).}\label{thm:topo safety}
\textit{\ref{sec:topo safety proof}
Let $f_\theta(x)$ be the SPM scalar generator using a strictly increasing control polygon. The mapping $f_\theta(x)$ is a global diffeomorphism immune to Runge's phenomenon, with a strictly positive first derivative $f_\theta'(x)$. Although strictly monotonic, the second derivative $f_\theta''(x)$ is topologically unconstrained. This allows it to form dynamic inflection points for Localized Spectral modelling}

\subsection{Instantiation I: Spectral-SPM (S-SPM) and eigenspace optimization}
Applying a scalar diffeomorphism to the eigenvalues of an SPD matrix creates a global diffeomorphism on the matrix space itself. Permutation invariance is important when handling unstructured physiological data with arbitrary sensor ordering. We define the S-SPM mapping $\phi_{S-SPM}: \mathcal{S}_{++}^n \to \mathcal{S}^n$ using the eigendecomposition $S = U \Sigma U^\top$:
\begin{equation}
\phi_{S-SPM}(S) = U \text{diag}(f_\theta(\lambda_1), \dots, f_\theta(\lambda_n)) U^\top
\end{equation}
\textbf{Theorem 3 (Diffeomorphism of S-SPM).} \label{thm:diff sspm}
\textit{ \ref{sec:diff sspm proof} The mapping $\phi_{S-SPM}: \mathcal{S}_{++}^n \to \mathcal{S}^n$ is a global diffeomorphism, and its isometric pullback induces a valid, bi-invariant Riemannian metric on $\mathcal{S}_{++}^n$.}

Integrating SPM into end-to-end deep learning architectures requires an analytically exact and numerically stable backward pass. Because computing the exact Lipschitz constant of a deep network is NP-hard \citep{scaman2018lipschitz}, network stability relies on ensuring that the operator norm of the local gradient at every layer is strictly bounded. We first state the analytical gradient for our mapping, followed by a mechanism to bound its Lipschitz constant.'

\textbf{Proposition 3 (Analytical Structural Gradient).} \label{thm:analytical}
\textit{\ref{sec:analytical proof} Let $X = \phi_{SPM}(S)$ be the forward mapping. Following the standard adjoint calculus for spectral functions \citep{ionescu2015matrix} and its unified closed-form derivation for SPD manifolds \citep{bhatia1997matrix}, the backpropagation gradient with respect to the input SPD matrix $S$ is analytically defined as:}
\begin{equation}
\nabla_S \mathcal{L} = U \left[ K \odot (U^\top (\nabla_X \mathcal{L}) U) \right] U^\top
\end{equation}
\textit{where $\mathcal{L}$ is the loss function, $\odot$ is the Hadamard product, and $K$ is the Daleck\u{\i}\u{\i}-Kre\u{\i}n divided difference matrix. Our monotonic spline $f_\theta$ constitutes a Primary Matrix Function \citep{bhatia1997matrix}, so $K$ evaluates using only the eigenvalue magnitudes:}
\begin{equation}
K_{ij} = 
\begin{cases} 
\frac{f_\theta(\lambda_i) - f_\theta(\lambda_j)}{\lambda_i - \lambda_j} & \text{if } \lambda_i \neq \lambda_j \\
f_\theta'(\lambda_i) & \text{if } \lambda_i = \lambda_j 
\end{cases}
\end{equation}

The orthogonal matrices $U$ hold an operator norm of $1$. The maximum absolute value in matrix $K$ bottlenecks the Lipschitz constant of this backward pass. When spectral singularities occur ($\lambda_i = \lambda_j$), evaluating $K_{ij}$ becomes unstable. We bound the operator norm using an asymmetric spectral perturbation.

\textbf{Theorem 4 (Gradient Bounding via Asymmetric Spectral Perturbation).}\label{thm:gradient}
\textit{\ref{sec:gradient proof} Let $\Sigma = \text{diag}(\lambda_1, \dots, \lambda_n)$ represent the sorted eigenvalue matrix from the spectral decomposition of $S$ ($\lambda_1 \le \lambda_2 \le \dots \le \lambda_n$). Before evaluating the Daleck\u{\i}\u{\i}-Kre\u{\i}n gradient matrix $K$, we perturb the eigenvalues in 64-bit precision:}
\begin{equation}
\tilde{\Sigma} = \Sigma_{f64} + \text{diag}(\vec{\epsilon})
\end{equation}
\textit{where $\vec{\epsilon} \in \mathbb{R}^n$ is an increasing vector such that $\epsilon_{i} = i \cdot \delta$ with a small machine constant $\delta > 0$. This asymmetric perturbation strictly bounds the Lipschitz constant of the backward pass.}

\subsection{Instantiation II: Cholesky-SPM (C-SPM) via product geometry}
For hierarchically ordered data, such as kinematic skeletons and radar arrays, the $\mathcal{O}(n^3)$ cost of eigendecomposition poses a computational bottleneck. C-SPM circumvents this by leveraging the intrinsic product geometry of the Cholesky manifold $\mathcal{L}_{++}^n$. Following recent advancements \citep{lin2019riemannian, chen2026fast}, the Cholesky manifold can be decoupled into a product space: $\mathcal{L}_{++}^n \cong \mathcal{SL}^n \times (\mathbb{R}_{>0})^n$, where $\mathcal{SL}^n$ is the flat Euclidean space of strictly lower triangular matrices, and $(\mathbb{R}_{>0})^n$ represents the strictly positive diagonals. 

We define a global diffeomorphism $\phi_{\mathcal{L}}$ that assigns the identity mapping to the Euclidean strictly lower triangular elements ($\lfloor \cdot \rfloor$), while applying our scalar diffeomorphism $f_\theta$ to the non-Euclidean diagonal elements:
\begin{equation}
\phi_{\mathcal{L}}(L) = \lfloor L \rfloor + \text{diag}\Big(f_\theta(L_{11}), \dots, f_\theta(L_{nn})\Big)
\end{equation}
The C-SPM metric on the target SPD manifold $\mathcal{S}_{++}^n$ is obtained via a composite pullback through the Cholesky decomposition $\text{Chol}: \mathcal{S}_{++}^n \to \mathcal{L}_{++}^n$. The composite mapping to the flat space $\mathcal{L}^n$ is therefore:
\begin{equation}
\phi_{C-SPM}(S) = \phi_{\mathcal{L}}(\text{Chol}(S))
\end{equation}

\textbf{Theorem 5 (Diffeomorphism of C-SPM).} \label{thm:diff cspm}
\textit{\ref{sec:diff cspm proof} The composite mapping $\phi_{C-SPM}: \mathcal{S}_{++}^n \to \mathcal{L}^n$ is a global diffeomorphism, and its pullback induces a valid Riemannian metric on $\mathcal{S}_{++}^n$.}

\textbf{Remark:} C-SPM avoids the Daleckĭĭ-Kreĭn structural gradient described in Proposition 2 by only working in the Cholesky product space. The backward pass depends entirely on the stable differentials of the Cholesky factorization, which gives it better native numerical stability and a speedup in algorithms.

\subsection{Geometric and topological properties of the SPM manifold}
\label{sec:spm_topology}
By establishing the Spline-Pullback Metric (SPM) through a global diffeomorphism to the ambient Euclidean space, the resultant Riemannian manifold acquires several advantageous topological characteristics that alleviate traditional computational challenges in SPD manifold learning.

\textbf{Theorem 6 (Global Flatness and Zero Sectional Curvature).} \label{thm:zero}
\textit{\ref{sec:zero proof} The Riemannian manifold $(\mathcal{S}_{++}^n, g_{SPM})$ induced by either the S-SPM or C-SPM pullback has constantly zero sectional curvature everywhere.}

\textbf{Theorem 7 (Existence, Uniqueness, and Closed-Form of the Fréchet Mean).}\label{thm:unique}
\textit{ \ref{sec:unique proof} For any finite set of SPD matrices $\{P_1, \dots, P_N\} \in \mathcal{S}_{++}^n$ with positive weights $w_i$ such that $\sum w_i = 1$, the SPM Fréchet Mean is globally unique and admits an exact closed-form expression.}

This eliminates the need for iterative Karcher flow approximations required by metrics with non-zero curvature. 

\textbf{Theorem 8 (Path-Independent Parallel Transport).} \label{thm:path}
\textit{\ref{sec:path proof} Let $A, B \in \mathcal{S}_{++}^n$ and $V \in \mathcal{T}_A\mathcal{S}_{++}^n$. The parallel transport of the tangent vector $V$ from $A$ to $B$ along the SPM geodesic is path-independent and evaluated in closed form.}

\textbf{Corollary 2 (Algebraic Structures).} \textit{Because $\phi_{S-SPM}$ and $\phi_{C-SPM}$ act as global isometries to flat vector spaces, the resulting metric space $\{\mathcal{S}_{++}^n, g_{SPM}\}$ inherits the algebraic structure of the ambient Euclidean space, forming an Abelian Lie group and a Hilbert space over $\mathbb{R}$.}\citep{arsigny2006log}

\textbf{Direct Implicit Projection:} Furthermore, because SPM is a highly parameterized global diffeomorphism, explicit Fréchet mean anchors for classification are mathematically redundant. By directly flattening the projection $v = \text{vec}(\phi(S))$, the B-spline dynamically learns to shift its vertical offset, implicitly centering the pullback space around the optimal Euclidean decision boundary. Because SPM acts as a global diffeomorphism to a flat Euclidean space, all fundamental Riemannian operators (Geodesic distance, Logarithm, Exponential maps) admit exact, closed-form expressions. The full derivations for these operators are provided in the Appendix \ref{sec:operators}

\section{Applications to deep Riemannian networks}
\label{sec:applications}

The closed-form and globally stable nature of SPM enables the robust construction of deep neural network architectures directly on the SPD manifold. We integrate SPM into two fundamental deep learning building blocks: Multinomial Logistic Regression and Residual Blocks.

\textbf{Riemannian Multinomial Logistic Regression (MLR):} Euclidean MLR acts as a hyperplane classifier \citep{lebanon2004hyperplane}, which generalizes to Riemannian manifolds via point-to-hyperplane distances: $p(y=k|X) \propto \exp(\langle A_k, \text{Log}_{P_k}(X) \rangle_{P_k})$, where $P_k$ is the class-specific manifold centroid and $A_k$ is the tangent normal vector\citep{chen2024rmlr}.

\textbf{Theorem 9 (SPM-MLR Classifier).}\label{thm:mlr}
\textit{ \ref{sec:mlr proof} Because the SPM framework operates as a global isometric pullback to a flat Euclidean space, the Riemannian inner product $\langle A_k, \text{Log}_{P_k}(X) \rangle_{P_k}$ algebraically simplifies to a standard Euclidean Frobenius inner product on the mapped matrices.}

This formulation eliminates the need to compute complex tangent space projections or inverse differentials during the final classification stage, maximizing computational throughput and stability.

\textbf{Riemannian Residual Blocks:} Deep Euclidean residual connections \citep{katsman2024riemannian} are generalized via the Riemannian Exponential map to ensure outputs remain on the SPD manifold: $X^{(i)} = \text{Exp}_{X^{(i-1)}}( \mathcal{F}_\omega(X^{(i-1)}) )$. Because SPM is geodesically complete, the network can process large residual tangent updates without violating topological constraints, unlike geodesically incomplete metrics.

\section{Numerical experiments }
\label{sec:numerical_experiments}

Before evaluating our proposed framework on real-world SPD manifold datasets, we conduct two controlled numerical experiments. These experiments are designed to empirically validate the topological safety bounds of Proposition 2 and to demonstrate the capacity limitations of classical metrics when faced with structured spectral noise.

\subsection{1D synthetic function approximation}
\begin{figure}[h]
    \centering
    \includegraphics[width=\linewidth]{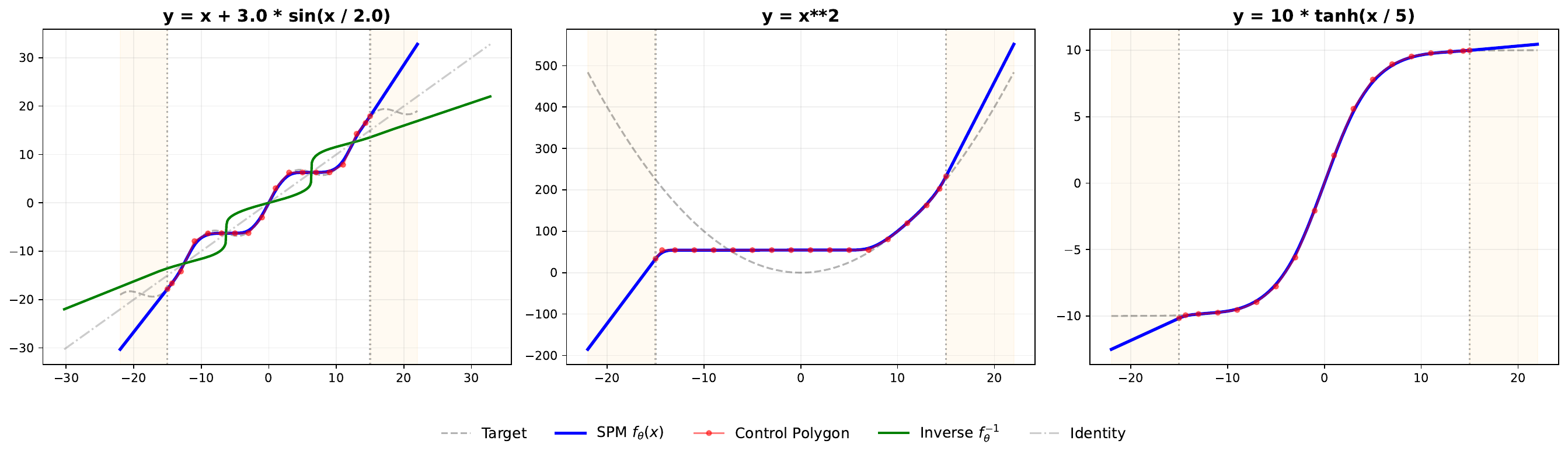} 
    \caption{Empirical validation of SPM properties. \textbf{(Left)}Unconstrained second-order differences ($\Delta^2 c_i$) enable the metric to closely monitor inflection points while maintaining monotonicity with the smooth inverse precluding singular gradients.  \textbf{(Center)} The SPM survives an adversarial, non-monotonic gradient by refusing to break bijectivity by maintaining a positive slope throughout.  \textbf{(Right)} SPM, utilizing its dynamic inflection capabilities, can cap noisy outliers, unachievable by fixed algebraic metrics.}
    \label{fig:1d_synthetic}
\end{figure}

We optimize the spline to approximate three different target functions, demonstrating its universal diffeomorphism approximation capabilities, as illustrated in Figure \ref{fig:1d_synthetic}. 

\subsection{Adversarial classification}
The defined monotonic derivatives of fixed geometries limit their ability to learn. To illustrate this, we create a synthetic binary classification dataset comprising 1000 $4 \times 4$ SPD matrices. The eigenspectrum is partitioned into four alternating bands: low-frequency noise $\lambda \in [0.1, 2.0]$, low-frequency signal, significant high-frequency noise $\lambda \in [12.0, 25.0]$, and high-frequency signal. The orthogonal matrices $U$ are chosen at random, which means that the linear probe can only use the learned spectral distances.

As detailed in Table \ref{tab:capacity_acc}, all non-adaptive classical baselines suffer from severe underfitting. Because their derivatives are fixed (such as $f'(\lambda) = 1/\lambda$ for LE), they face a trade-off: they must either over-amplify the low-frequency noise or over-amplify the high-frequency noise, blinding the classifier to the discriminative signal bands.

\begin{table}[htbp]
    \centering
    \caption{Classification accuracy on the alternating eigenvalue bands dataset. SPM successfully attains perfect performance, whereas rigid classical metrics lack the geometric capacity to separate the bands.}
    \label{tab:capacity_acc}
    \small 
    \setlength{\tabcolsep}{8pt} 
    \begin{tabular}{@{}llcc@{}} 
        \toprule
        \textbf{Metric Family} & \textbf{Geometry} & \textbf{Train Acc.} & \textbf{Test Acc.} \\ 
        \midrule
        \multirow{3}{*}{Rigid Baselines} & Log-Euclidean (LE) & $65.50\%$ & $66.00\%$ \\
        & Log-Cholesky (LC) & $63.50\%$ & $62.90\%$ \\
        & Power-Cholesky (PCM, $\theta=0.5$) & $64.50\%$ & $63.90\%$ \\ 
        \midrule
        \multirow{2}{*}{\textbf{Ours}} & C-SPM & $71.25\%$ & $69.90\%$ \\ 
        & \textbf{S-SPM} & \textbf{100.00\%} & \textbf{100.00\%} \\
        \bottomrule
    \end{tabular}
\end{table}

\begin{figure}[h]
    \centering
   
    \includegraphics[width=0.8\linewidth]{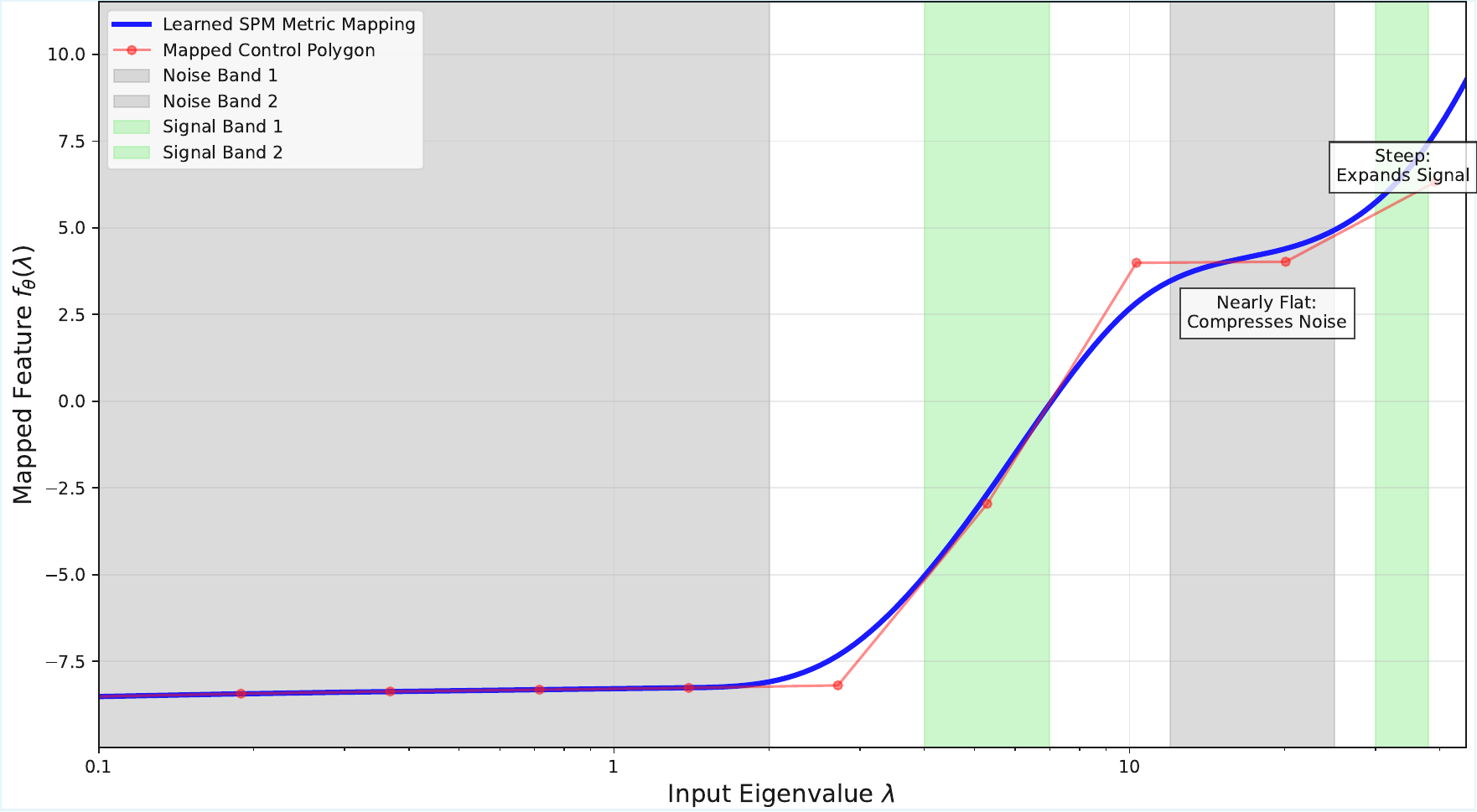} 
    \caption{S-SPM expressiviness and universal approximation capacity mapped over a logarithmic eigenspectrum. To achieve perfect class separability, the learned 
    s-SPM performs \textit{Localized Spectral Modelling}. It drives its derivative toward zero to horizontally compress the overlapping noise bands (gray) while simultaneously ramping vertically to stretch the discriminative signal bands (green). This sequence of dynamic inflections is mathematically impossible under classical metric priors.}
    \label{fig:alternating_bands}
\end{figure}

On the contrary, our proposed SPM achieves $100\%$ accuracy in both train and test sets, signalling the lack of overfitting. As visualized in Figure \ref{fig:alternating_bands}, SPM learns a staircase-function-like geometry to expand and minimize signal and noise, respectively. Moreover, the performance gap between SPM ($100\%$) and C-SPM ($69.90\%$) is attributable to the fact that the Cholesky decomposition mixes the eigenvalues with the randomized rotation matrix $U$. Hence, the success of C-SPM should be gauged by fairly comparing it with LC, over which it attains a prominent performance gain.  

\section{Experiments}
\label{sec:experiments}

To assess the state-of-the-art empirical efficacy of SPM, we use 3 widely used datasets, \textbf{HDM05\citep{muller2007mocap}}, \textbf{FPHA\citep{garcia2018first}}, and \textbf{Radar\citep{brooks2019riemannian}},   each with 3 different models, \textbf{Direct Linear Probe}, \textbf{SPDNet (SPD-MLR)}, and \textbf{Deep RResNet:}, across 7 metrics. The standard train-test split was used for FPHA evaluation, while 5-fold cross-validation was used for the rest. This set of datasets and models was chosen for a holistic evaluation and fair comparison with the literature due to its popular benchmarking usage in the literature \citep{chen2026fast, chen2024adaptive}. Further implementation details are presented in the appendix.

\begin{table}[htbp]
\centering
\caption{Classification accuracy on the HDM05 dataset. All models utilize 5-fold stratified cross-validation. Best results are highlighted in bold.}
\label{tab:hdm05}
\small 
\setlength{\tabcolsep}{10pt} 
\begin{tabular}{lccc}
\toprule
\textbf{Metric} & \textbf{Linear} & \textbf{SPD MLR} & \textbf{RResNet} \\ 
\midrule
AIRM & $0.5408 \pm 0.0413$ & $0.3876 \pm 0.0205$ & $0.5980 \pm 0.0244$ \\
LE & $0.5437 \pm 0.0441$ & $0.6594 \pm 0.0179$ & $0.6734 \pm 0.0227$ \\
LC & $0.5005 \pm 0.0260$ & $0.6066 \pm 0.0302$ & $0.6676 \pm 0.0209$ \\
\midrule
$\theta$-PCM ($\theta=0.5$) & $0.4962 \pm 0.0334$ & $0.6330 \pm 0.0168$ & $0.6580 \pm 0.0362$ \\
ALEM (Learn) & $0.5855 \pm 0.0568$ & $0.6719 \pm 0.0110$ & $0.6777 \pm 0.0195$ \\
\midrule
\textbf{C-SPM (Ours)} & $0.5831 \pm 0.0195$ & $0.6292 \pm 0.0306$ & $0.6498 \pm 0.0188$ \\
\textbf{S-SPM (Ours)} & \textbf{0.6032 $\pm$ 0.0341} & \textbf{0.6763 $\pm$ 0.0272} & \textbf{0.6801 $\pm$ 0.0124} \\
\bottomrule
\end{tabular}
\end{table}

\begin{table}[htbp]
\centering
\caption{Classification accuracy on the FPHA dataset. Evaluated on the standard predefined train-test split. }
\label{tab:fpha}
\small
\setlength{\tabcolsep}{10pt}
\begin{tabular}{lccc}
\toprule
\textbf{Metric} & \textbf{Linear} & \textbf{SPD MLR} & \textbf{RResNet} \\ 
\midrule
AIRM & $0.6226$ & $0.6174$ & $0.7461$ \\
LE & $0.8400$ & $0.8557$ & $0.8487$ \\
LC & $0.6957$ & $0.7426$ & $0.7739$ \\
\midrule
$\theta$-PCM & $0.6817$ \scriptsize{$(\theta=0.75)$} & $0.7357$ \scriptsize{$(\theta=0.75)$} & $0.7600$ \scriptsize{$(\theta=0.5)$} \\
ALEM (Learn) & $0.8539$ & $0.8852$ & $0.8487$ \\
\midrule
\textbf{C-SPM (Ours)} & $0.6887$ & $0.7391$ & $0.7791$ \\
\textbf{S-SPM (Ours)} & \textbf{0.8574} & \textbf{0.9043} & \textbf{0.8557} \\
\bottomrule
\end{tabular}
\end{table}

\begin{table}[htbp]
\centering
\caption{Classification accuracy on the Radar Signal Processing Dataset (5-fold CV). }
\label{tab:radar}
\resizebox{\columnwidth}{!}{%
\begin{tabular}{lccc}
\toprule
\textbf{Metric} & \textbf{Linear Probe} & \textbf{SPD MLR} & \textbf{RResNet} \\ 
\midrule
AIRM & $0.9627 \pm 0.0108$ & $0.9627 \pm 0.0100$ & $0.9087 \pm 0.0215$ \\
LE & $0.9630 \pm 0.0087$ & $0.9600 \pm 0.0129$ & $0.9267 \pm 0.0130$ \\
LC & $0.9677 \pm 0.0105$ & $0.9617 \pm 0.0115$ & $0.9303 \pm 0.0128$ \\
\midrule
$\theta$-PCM & $0.9660 \pm 0.0112$ \scriptsize{$(\theta=0.5)$} & $0.9593 \pm 0.0140$ \scriptsize{$(\theta=0.5)$} & $0.9183 \pm 0.0192$ \scriptsize{$(\theta=-1.0)$} \\
ALEM (Learnable) & $0.9650 \pm 0.0064$ & $0.9580 \pm 0.0090$ & $0.9210 \pm 0.0248$ \\
\midrule
\textbf{C-SPM (Ours)} & \textbf{0.9687 $\pm$ 0.0090} & $0.9617 \pm 0.0078$ & $0.9240 \pm 0.0177$ \\
\textbf{S-SPM (Ours)} & $0.9633 \pm 0.0075$ & \textbf{0.9657 $\pm$ 0.0076} & \textbf{0.9323 $\pm$ 0.0106} \\
\bottomrule
\end{tabular}%
}
\end{table}

 As demonstrated in Tables \ref{tab:hdm05}, \ref{tab:fpha}, and \ref{tab:radar}, the proposed SPM framework achieves state-of-the-art accuracy across all 9 architectural configurations. Predefined (LE/LCM) and scalar parametric metrics ($\theta$-PCM) rigidly induce information erasure during tangent space projection, failing to adapt when intermediate layers transform the manifold. Conversely, SPM securely maps these unregularized features. On the complex FPHA manifold, S-SPM ($0.9043$) provides a significant improvement over PCM ($0.7357$, Bootstrap $p<0.05$) in the SPDNet architecture, while also drastically outpacing baselines on uncompressed Linear Probes (like $0.8574$ vs. PCM's $0.6817$).

Finally, on the hierarchical Radar dataset, C-SPM achieves the highest linear probe accuracy ($0.9687$) and maintains highly competitive scores in deep RResNets. When strictly compared against other Cholesky-based metrics, C-SPM unveils a performance advantage while maintaining superior computational efficiency and numerical stability.

\section{Discussion and conclusion}
\label{sec:discussion}

The transition to SPM mirrors the paradigm shift from classical feature engineering to deep representation learning. While parametric metrics are fundamentally limited by their rigid mathematical priors, despite expensive grid searches, SPM, utilizing constrained B-splines, establishes a globally complete Euclidean pullback space and treats the Riemannian metric tensor itself as an end-to-end learnable representation, yielding state-of-the-art performance across a suite of experiments. As a parameter-efficient universal diffeomorphism approximator, SPM inherits the high expressivity of deep learning, opening three exciting frontiers in geometric research:

\textbf{Riemannian Metric Explainable AI (RMXAI):} While predefined metrics do not provide insight into the underlying data distribution, visualizing and interpreting the learned SPM spline offer direct insight into the dataset's spectral structure. We refer readers to Appendix \ref{sec:appendix_xai} for an RMXAI analysis of SPM and its stability.
    
\textbf{Riemannian Metric Initialization (RMI):} Like neural networks, SPM benefits from different initialization strategies. Future research in RMI will explore initialization distributions tailored to the SPM manifold, akin to Glorot\citep{glorot2010understanding} or He \citep{he2015delving} initialization, to ensure faster and more stable metric convergence.\\
\textbf{Riemannian Metric Regularization (RMR):}Though our experiments do not suffer from any pronounced overfitting, future applications of SPM in complex and data-scarce environments may lead to overfitting concerns. Work in RMR will find regularization techniques, beyond standard Euclidean weight decay, which would ensure the integrity of the topological and diffeomorphic rules of the manifold. A simple idea, for example, is to regularize the second derivative of the SPM curve.\\ An analysis of the framework's limitations is provided in Appendix \ref{sec:appendix_limitations}.

\bibliography{references}
\bibliographystyle{plainnat}

\newpage
\appendix

\section{Appendix: preliminaries}
\textbf{Riemannian Calculus and Operators}

Endowing the topological space $\mathcal{S}_{++}^n$ with a Riemannian metric $g$ allows for the generalization of Euclidean calculus to the manifold. For any points $P, Q \in \mathcal{S}_{++}^n$ and a tangent vector $V \in \mathcal{T}_P\mathcal{S}_{++}^n$, the fundamental Riemannian operators are defined as follows:

\textbf{Geodesics and Distance:} A geodesic $\gamma(t)$ is the locally length-minimizing curve connecting two points on the manifold, acting as the non-Euclidean generalization of a straight line. The geodesic distance $d_g(P, Q)$ is the integrated length of this curve under the metric $g$.

\textbf{Exponential and Logarithmic Maps:} To traverse the manifold, one must map between the curved space and its flat tangent approximations. The Riemannian Exponential map, $\text{Exp}_P(V): \mathcal{T}_P\mathcal{S}_{++}^n \to \mathcal{S}_{++}^n$, projects a tangent vector back onto the manifold such that the curve $\gamma(t) = \text{Exp}_P(tV)$ is a geodesic with initial velocity $V$. Its inverse, the Riemannian Logarithm map, $\text{Log}_P(Q): \mathcal{S}_{++}^n \to \mathcal{T}_P\mathcal{S}_{++}^n$, calculates the exact tangent vector (initial velocity) required to reach $Q$ from $P$.

\textbf{The Fréchet Mean:} Because arithmetic averaging falls outside the SPD cone, the center of mass for a set of points $\{P_i\}_{i=1}^N$ with weights $w_i$ must be computed geometrically. The Riemannian Fréchet Mean generalizes the Euclidean average by minimizing the sum of squared geodesic distances\citep{karcher1977riemannian}:
\begin{equation}
    \mu_F = \arg\min_{M \in \mathcal{S}_{++}^n} \sum_{i=1}^N w_i d_g^2(M, P_i)
\end{equation}

Under metrics with non-zero sectional curvature (e.g., AIRM), computing the Fréchet mean requires iterative, computationally expensive gradient descent (Karcher flow). Conversely, under the globally flat geometry of a Pullback Euclidean Metric, the Fréchet mean and all associated Riemannian maps admit mathematically exact, closed-form Euclidean solutions within the pullback space.

\textbf{Splines in Geometric Computing.} 
While B-splines have a rich history in computational anatomy (like LDDMM \citep{beg2005computing}) and image registration (like  Free-Form Deformations\citep{rueckert1999nonrigid}), these classical frameworks utilize splines to parameterize \textit{spatial} coordinate grids to align volumetric data, and recently, the benefits of their Universal Approximation Properties in deep learning are gaining traction, as evidenced by their utilization in Kolmogorov-Arnold Networks (KANs) \citep{liu2024kan}. Furthermore, in SPD literature, splines are occasionally used for temporal trajectory interpolation \textit{on top of} a fixed, predefined Riemannian metric. However, for the first time, instead of using splines to traverse a manifold spatially or temporally, the SPM framework uses learnable, strictly monotonic B-splines to construct the spectral Riemannian metric tensor itself, reducing complex multi-dimensional Jacobian constraints to a computationally trivial 1D monotonicity constraint.

\section{Proofs of theoretical claims}
\label{sec:proofs}

\subsection{Proof of theorem \hyperref[thm:global scalar diff]{1}}
\label{sec:global scalar diff proof}
\textit{Proof.} A smooth, globally bijective map with a smooth inverse forms a global diffeomorphism\citep[Ch.~2, p.~39]{lee2012smooth}. Our scalar generator is defined as a composition: $f_\theta(x) = S(\log(x))$, where $\log: \mathbb{R}_{>0} \to \mathbb{R}$ is the standard natural logarithm, and $S: \mathbb{R} \to \mathbb{R}$ is the monotonic B-spline with linear boundary extrapolation. The natural logarithm is a standard global $C^\infty$ diffeomorphism. Since composing two diffeomorphisms yields another diffeomorphism \citep[Ch.~2, Prop.~2.15]{lee2012smooth}, we can prove $f_\theta(x)$ meets the necessary topological requirements. We verify the specific criteria below:

\textbf{1. Global Bijection (One-to-One and Onto):}Let $y = \log(x)$ be the input to the spline. Inside the knot vector $t_k \le y \le t_{m-k}$, the B-spline basis functions possess strictly local support and are globally non-negative ($B_{i,k-1}(y) \ge 0$) \citep[Ch.~IX, Prop.~ii]{deboor2001}. The partition of unity property ($\sum B_{i,k-1}(y) = 1$) \citep[Ch.~IX, Prop.~iv]{deboor2001} guarantees the strict positivity ($>0$) of at least one basis function at any given $y$. The knot increments are strictly positive. Hence, according to Proposition 1, the sign of the spline derivative $S'(y)$ depends only on the coefficient differences $c_i - c_{i-1}$. Under our architectural constraint $c_i > c_{i-1}$, it mathematically follows that $S'(y) > 0$. Outside the knot vector, the boundary slopes $m_L$ and $m_R$  match the strictly positive derivatives at the grid edges, ensuring $S'(y) > 0$ globally on $\mathbb{R}$. We evaluate the total derivative of the composite function with respect to the original eigenvalue $x \in \mathbb{R}_{>0}$ via the chain rule, which yields $f_\theta'(x) = S'(\log x) \cdot \frac{1}{x}$. Because $x > 0$ and $S'(y) > 0$, the gradient $f_\theta'(x) > 0$ is strictly bounded above zero, rendering the function strictly monotonically increasing and thus injective (one-to-one). Regarding surjectivity, the logarithmic mapping pushes the manifold boundary ($x \to 0^+$) to $y \to -\infty$. Because the linear extrapolation extends to infinity on both tails with strictly positive slopes, $S(y) \to \pm\infty$ as $y \to \pm\infty$. The range covers the entire unconstrained real line $(-\infty, \infty)$, rendering it surjective (onto). Thus, $f_\theta(x)$ is a global bijection.

\textbf{2. Smooth Forward Map ($C^1$ Continuity):}Standard B-splines of degree $k \ge 3$ are at least $C^2$ continuous inside their knot grid. We clamp the terminal knots of the grid with multiplicity $k$ to transition smoothly to the unconstrained domain. Following standard spline behavior for repeated knots\citep[Ch.~IX]{deboor2001}, this multi-knot insertion intentionally reduces continuity at the edges, forcing the curve to interpolate the boundary control points directly. Consequently, the boundary derivatives are defined by the first and last segments of the control polygon, which are adopted by the linear extrapolation, at the boundaries $t_k$ and $t_{m-k}$, to ensure a globally smooth forward mapping and a $C^1$ continuity at the transition points, by matching the tangent slopes $m_L$ and $m_R$. 

\textbf{3. Smooth Inverse Map:} The mapping requires a smooth inverse to be a diffeomorphism. By the Inverse Function Theorem \citep[Appendix~C, Thm.~C.34]{lee2012smooth}, if a continuously differentiable mapping has an invertible total derivative (Jacobian) at every point, a $C^1$ smooth inverse is guaranteed to exist. For a scalar function $f_\theta: \mathbb{R}_{>0} \to \mathbb{R}$, the Jacobian is simply a $1 \times 1$ matrix $[f_\theta'(x)]$. The monotonic parameterization and the chain rule keep $f_\theta'(x) > 0$ globally, rendering the determinant non-zero and satisfying the invertibility condition. Consequently, $f_\theta(x)$ is a smooth bijection with a smooth inverse, forming a true global diffeomorphism \citep[Ch.~2, p.~39]{lee2012smooth}.

\textbf{Conclusion:}Because $f_\theta(x)$ is the composition of the global diffeomorphism $\log(x)$ and the global $C^1$ diffeomorphism $S(y)$, it mathematically constitutes a global $C^1$ diffeomorphism from $\mathbb{R}_{>0}$ to $\mathbb{R}$. $\hfill \blacksquare$

\subsection{Proof of theorem \hyperref[thm:univ approx]{2}}
\label{sec:univ approx proof}
\textit{Proof.} Let $g \in \text{Diff}_+^1(\mathbb{R})$ be an arbitrary target strictly monotonic $C^1$ diffeomorphism, and let $[a,b]$ be the compact interval covering the dataset's logarithmic spectrum. The Schoenberg variation-diminishing property of B-splines allows Schoenberg's spline operator to uniformly approximate $g(x)$ on $[a,b]$ \citep[Ch.~XI, Eq.~33]{deboor2001}. This yields a spline $S_k(x) = \sum c_i B_{i,k}(x)$ over a finite knot vector $T$ covering $[a,b]$.

If the control coefficients are chosen as $c_i = g(\xi_i)$, where $\xi_i$ are the Greville abscissae (knot averages \citep[Ch.~IX, Eq.~39; Ch.~XI, Eq.~33]{deboor2001}), the spline uniformly converges to $g(x)$ as the knot spacing $h \to 0$. Since $g(x)$ strictly increases, any $\xi_i > \xi_{i-1}$ means the sampled control points satisfy $g(\xi_i) > g(\xi_{i-1})$. As established by \citet[Ch.~XI]{deboor2001}, this shape-preserving operator maps monotone functions to monotone functions, which yields our architectural constraint $c_i > c_{i-1}$. (Note: In the discrete neural instantiation, the parameter $\epsilon$ in $c_i = c_{i-1} + \text{softplus}(w_i) + \epsilon$ acts as a strictly positive lower bound. For true asymptotic universal approximation as $h \to 0$, it is required that $\epsilon \to 0$ such that $\epsilon < \min_i (g(\xi_i) - g(\xi_{i-1}))$.)

We use cubic B-splines (order $=4$), and the target $g(x)$ is $C^1$ smooth. Under Schoenberg's operator, the uniform convergence applies simultaneously to the first derivative \citep{Marsden1970, Gonska2010}: $\sup_{x \in [a,b]} |S_k'(x) - g'(x)| \to 0$ as $h \to 0$.

The linear extrapolation of $f_\theta(x)$ maintains global topological bijectivity outside $T$. Inside $[a,b]$, the spline holds dense $C^1$ approximation capacity. The function $f_\theta$ and its derivative $f_\theta'$ entirely parameterize the SPM pullback metric tensor and its Daleck\u{\i}\u{\i}-Kre\u{\i}n structural gradient. This simultaneous $C^1$ convergence ensures SPM universally approximates the optimal Riemannian metric tensor for a given dataset. $\hfill \blacksquare$

\subsection{Proof of corollary \hyperref[cor:univ subsump]{1}}
\label{sec:univ subsump proof}
\textit{Proof.} Let $y = \log(x)$ be the input to the parameterized spline $S(y)$. Because standard backpropagation relies on the first derivative to compute the Daleck\u{\i}\u{\i}-Kre\u{\i}n gradient, the $C^1$ convergence guaranteed by Theorem 2 provides the necessary and sufficient topological smoothness to functionally subsume these target metrics for deep learning optimization. We review the functional mapping $S(y)$ needed for SPM to recover existing geometries:

\textbf{1. Logarithmic Metrics (LE, LC):} These use the standard natural logarithm $f(x) = \log(x)$. Equating this to the SPM composition gives $S(\log x) = \log x$, simplifying to $S(y) = y$. Hence, SPM can easily replicate these by learning the identity function across the control polygon.

\textbf{2. Power-Cholesky Metric (PCM):} Parametric power metrics rely on fractional powers $f(x) = x^\theta$ for $\theta > 0$. The SPM composition becomes $S(\log x) = x^\theta$. Substituting $x = e^y$, the required spline mapping is $S(y) = e^{\theta y}$. The exponential function $e^{\theta y}$ is a strictly increasing $C^\infty$ diffeomorphism. Theorem 2 ensures the monotonic B-spline can uniformly approximate the PCM geometry over any compact spectral domain.

An analysis of ALEM is not presented due to its diffeomorphic limitations.

\textbf{Conclusion:}As a universal approximator for diffeomorphisms, the Spline-Pullback Metric acts as a  superset of existing analytical metrics. $\hfill \blacksquare$

\subsection{Proof of proposition \hyperref[thm:topo safety]{2}}
\label{sec:topo safety proof}
\textit{Proof.} First, we evaluate topological safety. Learning a non-linear metric mapping via high-degree polynomial regression usually triggers Runge's phenomenon. This generates macro-oscillations at the boundaries, forcing the first derivative to cross zero ($f_\theta'(x) \le 0$) and breaking the bijection. The SPM generator avoids this by applying Schoenberg's Variation-Diminishing operator to cubic B-splines \citep[Ch.~XI, Cor.~28]{deboor2001}. The parameterization requires a monotonic control polygon ($c_i > c_{i-1}$). The variation-diminishing property then stops the continuous curve $f_\theta(x)$ from oscillating. We clamp the boundary knots to interpolate the terminal control points without overshoot. This yields a global diffeomorphism safe from Runge's phenomenon.

Next, we establish geometric expressivity. Let $\Delta c_i = c_i - c_{i-1}$ denote the first-order finite difference of the control points. To ensure monotonicity, we parameterize the control polygon via a cumulative sum of strictly positive steps: $c_i = c_0 + \sum_{j=1}^i \Delta c_j$. The steps themselves are parameterized as unconstrained neural network weights processed through a strictly positive activation, $\Delta c_j = \text{softplus}(w_j) + \epsilon$. According to Proposition 1, enforcing $\Delta c_i > 0$ bounds the first derivative strictly above zero ($f_\theta'(x) > 0$).

However, the second-order finite difference, $\Delta^2 c_i = \Delta c_{i+1} - \Delta c_i$, remains unconstrained ($\Delta^2 c_i \in \mathbb{R}$), as the underlying weights $w_j$ can be optimized to any real value. Because the control points of the first derivative spline curve are proportional to $\Delta c_i$, the control polygon of this derivative curve, though strictly above the x-axis, is free to undergo sign changes with respect to any line $y=\lambda, \lambda>0$, giving freedom to the underlying spline curve to do so \citep[Ch.~XI, Cor.~28]{deboor2001}. This allows the second spline derivative $f_\theta''(x)$ to cross zero and dynamically form inflection points. Unlike classical metrics such as the Log-Euclidean ($f''(x) = -1/x^2$) or Power-Cholesky, which impose a fixed global concavity or convexity, the unconstrained $\Delta^2 c_i$ in SPM enables Localized Spectral modelling. The metric can learn "S-shaped" deformations that compress noise and expand specific eigenvalue subspaces. This flexibility is impossible with rigid algebraic priors $\hfill \blacksquare$

\subsection{Proof of theorem \hyperref[thm:diff sspm]{3}}
\label{sec:diff sspm proof}
\textit{Proof.}According to Theorem 1, the scalar generator $f_\theta(x)$ constitutes a $C^1$ global diffeomorphism from $\mathbb{R}_{>0}$ to $\mathbb{R}$. We elevate this scalar mapping to the matrix manifold via the calculus of Primary Matrix Functions, defining the forward mapping $\phi_{S-SPM}(S) = U f_\theta(\Sigma) U^\top$ \citep[Ch.~V.1]{bhatia1997matrix}.

\textbf{1. Matrix Bijection:} Because $f_\theta$ is a  scalar bijection, its inverse $f_\theta^{-1}: \mathbb{R} \to \mathbb{R}_{>0}$ uniquely exists. By the same calculus, this induces a well-defined inverse matrix function $\phi^{-1}(Y) = U f_\theta^{-1}(\Sigma_Y) U^\top$. Applying this inverse yields the identity mappings $\phi^{-1}(\phi(X)) = X$ and $\phi(\phi^{-1}(Y)) = Y$. The first identity ensures injectivity (as $\phi(X_1) = \phi(X_2) \implies X_1 = X_2$). The second ensures surjectivity (as every $Y$ has a valid pre-image), without repeated eigenvalues posing any challenge, as observed in rank-based metrics. This two-sided inverse guarantees that the matrix mapping $\phi_{S-SPM}$ is a strict global bijection between the SPD manifold $\mathcal{S}_{++}^n$ and the space of symmetric matrices $\mathcal{S}^n$.

\textbf{2. Matrix Smoothness:} The Daleck\u{\i}\u{\i}-Kre\u{\i}n theorem \citep[Thm.~V.3.3]{bhatia1997matrix} states that applying a $C^1$ smooth scalar function to a symmetric matrix creates a Primary Matrix Function that is continuously Fréchet differentiable. Both $f_\theta$ and $f_\theta^{-1}$ are $C^1$ smooth scalars. As a result, the forward matrix mapping $\phi$ and its inverse $\phi^{-1}$ are globally $C^1$ smooth. The matrix mapping inherits the scalar generator's strict bijection and smooth invertibility. This meets all criteria for a global diffeomorphism and induces a valid bi-invariant Riemannian metric.

\textbf{3. Bi-invariance:} 
Because the forward mapping $\phi_{S-SPM}(S)$ is defined via the spectral decomposition, it constitutes a Primary Matrix Function \citep[Ch.~V]{bhatia1997matrix}. By definition, such functions commute with orthogonal congruence transformations ($\phi(RSR^\top) = R\phi(S)R^\top$ for any orthogonal $R$). Consequently, pulling back the orthogonally invariant Frobenius inner product from the ambient space of symmetric matrices via the Riemannian isometry guarantees that the induced S-SPM geometry is  bi-invariant. $\hfill \blacksquare$

\subsection{Proof of proposition \hyperref[thm:analytical]{3}}
\label{sec:analytical proof}
\textit{Proof.} This closed-form gradient is a direct consequence of applying the Fréchet differential of symmetric matrix functions \citep[Thm.~V.3.3]{bhatia1997matrix} to the Frobenius inner product chain rule $d\mathcal{L} = \langle \nabla_X \mathcal{L}, dX \rangle$. By substituting the analytical B-spline basis evaluations for $f_\theta$ and $f_\theta'$ into the divided difference matrix $K$, we obtain the structural gradient required for Riemannian backpropagation. $\hfill \blacksquare$

\subsection{Proof of theorem \hyperref[thm:gradient]{4}}
\label{sec:gradient proof}
\textit{Proof.} Let the perturbed eigenvalues be $\tilde{\lambda}_i = \lambda_i + \epsilon_i$. Because the original eigenvalues are sorted (generally intrinsically by computational eigenvalue solvers) ($\lambda_{i+1} - \lambda_i \ge 0$), and $\vec{\epsilon}$ enforces a strict minimal growth $\epsilon_{i+1} - \epsilon_i = \delta > 0$, we mathematically guarantee that for all $i < j$:
\begin{equation}
\tilde{\lambda}_j - \tilde{\lambda}_i = (\lambda_j - \lambda_i) + (\epsilon_j - \epsilon_i) \ge \delta > 0
\end{equation}
This establishes a minimal spectral gap $|\tilde{\lambda}_i - \tilde{\lambda}_j| \ge \delta$ for all $i \neq j$. Applying this minimum gap to the Daleck\u{\i}\u{\i}-Kre\u{\i}n matrix $K$, the off-diagonal elements are strictly bounded:
\begin{equation}
\max_{i \neq j} |K_{ij}| \le \frac{|f_\theta(\tilde{\lambda}_i) - f_\theta(\tilde{\lambda}_j)|}{\delta}
\end{equation}
The B-spline output $f_\theta$ is continuously differentiable and bounded smoothly by its control points, meaning the numerator remains finite. The denominator is strictly positive and bounded below by $\delta$. Together, this permanently bounds the gradient norm. Capping the Lipschitz constant of the backward pass this way protects the Riemannian optimization from eigenvalue degeneracy and $0/0$ gradient explosions. As $\delta \to 0$, the perturbed off-diagonal elements in the matrix naturally converge to the true analytical derivative $f_\theta'(\lambda)$ through the finite-difference limit.$\hfill \blacksquare$

\subsection{Proof of theorem \hyperref[thm:diff cspm]{5}}
\label{sec:diff cspm proof}
\textit{Proof.} We establish this hierarchically through the product space. The mapping $\phi_{\mathcal{L}}$ acts as the identity on the Euclidean subspace $\mathcal{SL}^n$ and as the scalar diffeomorphism $f_\theta$ (established in Theorem 1) on the diagonal subspace $(\mathbb{R}_{>0})^n$. Because both subspace mappings are $C^1$ diffeomorphisms, their direct product $\phi_{\mathcal{L}}$ is a global diffeomorphism from $\mathcal{L}_{++}^n$ to $\mathcal{L}^n$. Furthermore, the standard Cholesky factorization $\text{Chol}: \mathcal{S}_{++}^n \to \mathcal{L}_{++}^n$ is a well-established global diffeomorphism \citep{lin2019riemannian}. Because the composition of two diffeomorphisms is itself a diffeomorphism \citep[Ch.~2, Prop.~2.15]{lee2012smooth}, $\phi_{C-SPM} = \phi_{\mathcal{L}} \circ \text{Chol}$ forms a global bijection from $\mathcal{S}_{++}^n$ to the Euclidean space $\mathcal{L}^n$ with a smooth inverse, establishing a  valid pullback geometry. $\hfill \blacksquare$

\subsection{Proof of theorem \hyperref[thm:zero]{6}}
\label{sec:zero proof}
\textit{Proof.} Let $\phi: \mathcal{S}_{++}^n \to \mathcal{E}$ be the SPM global diffeomorphism mapping the SPD manifold to the Euclidean vector space $\mathcal{E}$ (where $\mathcal{E} = \mathcal{S}^n$ for S-SPM, and $\mathcal{E} = \mathcal{L}^n$ for C-SPM). The target space $\mathcal{E}$ is an unconstrained flat vector space because our scalar generator $f_\theta$ is globally surjective onto $\mathbb{R}$. The Christoffel symbols of the Levi-Civita connection vanish in the transformed coordinates because $g_{SPM}$ is defined as the isometric pullback of the standard Euclidean inner product $\langle \cdot, \cdot \rangle_F$ through $\phi$.  As a result, the Riemannian curvature tensor becomes zero globally, rendering the manifold globally flat (analogous to the isometric preservation of zero sectional curvature established in Log-Euclidean and Log-Cholesky Spaces \citep{arsigny2006log,lin2019riemannian} while upgrading the isometry to a learnable parameterization. $\hfill \blacksquare$

\subsection{Proof of theorem \hyperref[thm:unique]{7}}
\label{sec:unique proof}
\textit{Proof.} The Fréchet Mean on a Riemannian manifold is defined as the minimizer of the sum of squared geodesic distances: $P^* = \arg\min_{P \in \mathcal{S}_{++}^n} \sum w_i d_{SPM}^2(P, P_i)$. Because the manifold $(\mathcal{S}_{++}^n, g_{SPM})$ has zero sectional curvature (Theorem 5) and is simply connected, the Fréchet variance function is strictly convex\citep{karcher1977riemannian}. Therefore, the unique global minimum exists and maps identically to the weighted Euclidean mean in the pullback space \citep{pennec2006riemannian}:
\begin{equation}
    \text{WFM}_{SPM}(\{P_i\}, \{w_i\}) = \phi^{-1} \left( \sum_{i=1}^N w_i \phi(P_i) \right)
\end{equation}
$\hfill \blacksquare$

\subsection{Proof of theorem \hyperref[thm:path]{8}}
\label{sec:path proof}
\textit{Proof.} In a manifold with zero sectional curvature, parallel transport between any two points is independent of the curve connecting them\citep{lee2018introduction}. Under the SPM isometry, parallel transport equates to projecting the tangent vector $V$ to the flat Euclidean space at $A$, translating it to $B$ (which acts as the identity operation in a flat vector space), and projecting it back to the tangent space at $B$:
\begin{equation}
    \text{PT}_{A \to B}(V) = \mathcal{D}\phi_B^{-1} \Big( \mathcal{D}\phi_A(V) \Big)
\end{equation}
where $\mathcal{D}\phi$ and $\mathcal{D}\phi^{-1}$ are the forward and inverse pushforward differentials of the SPM mapping. $\hfill \blacksquare$

\subsection{Proof of theorem \hyperref[thm:mlr]{9}}
\label{sec:mlr proof}

\textit{Proof.} Let $\tilde{A}_k$ represent the unconstrained Euclidean weights of the classifier corresponding to the tangent vector $A_k$, and let $\phi \in \{\phi_{S-SPM}, \phi_{C-SPM}\}$. By substituting the closed-form SPM Riemannian Logarithm into the MLR probability function, the Riemannian geometry collapses into a flat inner product. Given an input $X \in \mathcal{S}_{++}^n$, the SPM-MLR is evaluated as:
\begin{equation}
    p(y=k | X) \propto \exp \Big( \langle \tilde{A}_k, \phi(X) - \phi(P_k) \rangle_F \Big)
\end{equation}$\hfill \blacksquare$

\section{Riemannian and algebraic operators on the SPM manifold}
\label{sec:operators}

The Spline-Pullback Metric framework functions as a global diffeomorphism, mapping the curved manifold $\mathcal{S}_{++}^n$ to flat Euclidean spaces, the space of symmetric matrices $\mathcal{S}^n$ for S-SPM, and lower triangular matrices $\mathcal{L}^n$ for C-SPM. Consequently, all fundamental Riemannian and algebraic operators can be derived in closed form through the pullback mechanism.

\subsection{Differentials and tangent projections}
For an SPD matrix $P \in \mathcal{S}_{++}^n$, let $P = U \Sigma U^\top$ be its eigendecomposition and $P = L L^\top$ be its Cholesky factorization. The differential map projecting a tangent vector $V \in \mathcal{T}_P\mathcal{S}_{++}^n$ to the respective representation spaces depends on the chosen SPM instantiation:
\begin{enumerate}
    \item \textbf{Spectral Differential (S-SPM):} $\mathcal{D}_{eig}(V) = U [ K \odot (U^\top V U) ] U^\top$, where $K$ is the Daleckĭĭ-Kreĭn matrix derived in Proposition 2.
    \item \textbf{Cholesky Differential (C-SPM):} $\mathcal{D}_{chol}(V) = \lfloor L(L^{-1} V L^{-\top}) \rfloor + \frac{1}{2}\text{diag}(L^{-1} V L^{-\top})L$, adhering to standard Cholesky pushforward differentials \citep{lin2019riemannian}.
\end{enumerate}

\subsection{Closed-form Riemannian operators}
Let $\phi \in \{\phi_{S-SPM}, \phi_{C-SPM}\}$ denote the chosen global diffeomorphism, and let $\mathcal{D}_{\phi}$ denote its respective differential map. For any $P, Q \in \mathcal{S}_{++}^n$ and tangent vector $V \in \mathcal{T}_P\mathcal{S}_{++}^n$, the globally complete Riemannian operators are:
\begin{align}
    \text{Geodesic Path ($t \in [0,1]$):} \quad & \gamma_{SPM}(P, Q; t) = \phi^{-1} \Big( (1-t)\phi(P) + t\phi(Q) \Big) \\
    \text{Riemannian Logarithm:} \quad & \text{Log}_P(Q) = \mathcal{D}_{\phi}^{-1} \Big( \phi(Q) - \phi(P) \Big) \\
    \text{Riemannian Exponential:} \quad & \text{Exp}_P(V) = \phi^{-1} \Big( \phi(P) + \mathcal{D}_{\phi}(V) \Big) \\
    \text{Geodesic Distance:} \quad & d_{SPM}^2(P, Q) = \| \phi(P) - \phi(Q) \|_F^2
\end{align}

\section{Riemannian Metric Explainable AI (RMXAI)}
\label{sec:appendix_xai}

A central requirement for Riemannian Metric Explainable AI (RMXAI) is that the learned geometric deformations represent stable and generalizable data characteristics. In order to empirically verify the stability of the proposed method, the final learned Spectral-SPM (S-SPM) diffeomorphisms were recorded in all 5 independent folds of the Stratified Cross-Validation on the Radar dataset.

\begin{figure}[h]
    \centering
   
    \includegraphics[width=\linewidth]{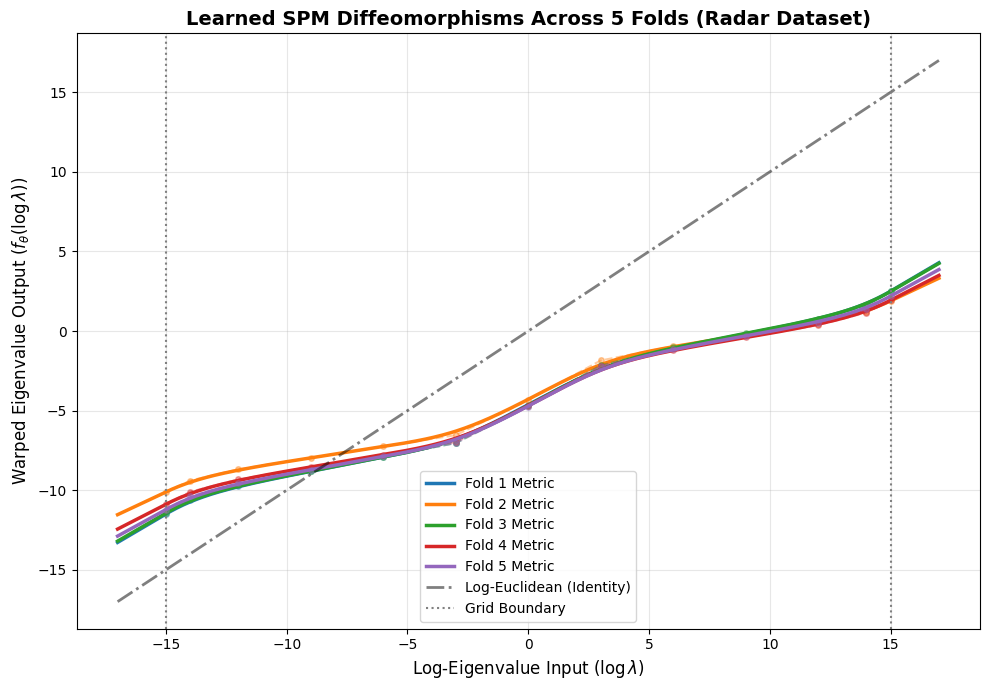} 
    \caption{Learned Spectral-SPM diffeomorphisms across 5-fold cross-validation on the Radar signal processing dataset. The consistent convergence to a non-linear topology across independent folds highlights the stability of the SPM optimization and its ability to act as an interpretable diagnostic tool for spectral noise.}
    \label{fig:5fold_radar}
\end{figure}

As illustrated in Figure \ref{fig:5fold_radar}, despite being exposed to different data splits, the monotonic B-splines converged to consistent topological shapes, learning to compress noisy logarithmic eigenvalues.  This demonstrates a stable geometric grouping across the folds, indicating that SPM optimization is stable and effectively captures the inherent spectral properties of the dataset.

\section{Network architectures and implementation}
\label{sec:appendix_c}

\subsection{Optimization and training protocols}
Across all datasets and architectures, we maintained a uniform training protocol to prevent hyperparameter overfitting. To eliminate variance from stochasticity, a universal random seed (\texttt{seed=42}) was enforced across all hardware and data-loading workers. The PyTorch library\citep{paszke2019pytorch} was used for the implementations. All experiments were conducted on NVIDIA GPU T4 hardware. The full source code and complete training scripts will be publicly released on GitHub upon the acceptance of this manuscript. Comprehensive hyperparameter details and architectural configurations are provided below to ensure theoretical and algorithmic reproducibility:
\begin{itemize}
    \item \textbf{Optimizer:} Riemannian Adam (\texttt{geoopt.optim.RiemannianAdam}) 
    \item \textbf{Hyperparameters:} Learning rate $lr = 10^{-3}$, weight decay $= 10^{-4}$, and batch size $= 32$.
    \item \textbf{Gradient Norm:} For stable convergence, a maximum gradient norm of $2.0$ was implemented.
    \item \textbf{Early Stopping:} Models were trained for a maximum of $100$ epochs. We utilized an early stopping criterion with a patience of $15$ epochs, monitored via the accuracy on a $20\%$ internal validation split isolated from the training fold. The weights from the best-performing validation epoch were restored for final test-set inference.
\end{itemize}

\subsection{ Numerical safeguards }
Implementation Detail: While Theorem 4 establishes the theoretical Lipschitz bounds via direct spectral perturbation, standard automatic differentiation frameworks compute the Fréchet derivative internally and will fault on exact degeneracies before the gradient matrix can be manually bounded. Therefore, in our hardware implementation, we apply a functionally equivalent symmetry-breaking diagonal perturbation to the input matrix $S$ in 64-bit precision immediately prior to the eigensolver, to ensure numerical stability during the backward pass.

As noted in the main text, full-rank kinematic SPD matrices (e.g., HDM05) are severely rank-deficient and frequently precipitate \texttt{LinAlgError} crashes during standard 32-bit eigendecomposition or Cholesky factorization. All models in our suite (including our stabilized implementations of the baselines) were equipped with the following numerical armor:
\begin{itemize}
    \item \textbf{Symmetry Enforcement \& 64-Bit Upcasting:} Prior to any decomposition, matrices were strictly symmetrized ($X = \frac{1}{2}(X + X^\top)$) and upcast to \texttt{Float64} precision.
    \item \textbf{Eigenvalue Jitter (Tie-Breaking):} To prevent degenerate eigenvalue collapse, a uniformly spaced infinitesimal tie-breaker ($\texttt{torch.linspace}(10^{-8}, 2 \cdot 10^{-8}, n)$) was added to the diagonal prior to decomposition.
    \item \textbf{Cholesky Fallback Mechanism:} For Cholesky-based metrics, we implemented a dual-stage factorization. The primary attempt utilized a standard $10^{-4}I$ jitter. If the PyTorch backend threw a \texttt{LinAlgError}, the module dynamically intercepted the exception, fell back to an eigendecomposition, hard-clamped all eigenvalues to a minimum of $10^{-4}$, reconstructed the matrix, and successfully performed the Cholesky factorization on the sanitized matrix.
\end{itemize}

\subsection{ Spline-Pullback Metric (SPM) parameters}
The SPM mapping was parameterized using a Cox-de Boor B-spline evaluated on a clamped knot vector.

\begin{itemize}
    \item \textbf{Grid Configuration:} $\text{Grid Size} = 10$, $\text{Spline Order} = 3$ (cubic), $\text{Domain Range} = [-15.0, 15.0]$. To ensure a fair and controlled comparison, we fix this spline configuration across all models and datasets for both S-SPM and C-SPM.
    \item \textbf{Monotonicity Constraint:} The control points $c_i$ were dynamically generated via cumulative summation of softplus-activated weights, ensuring $C^1$ monotonic bijectivity.
    \item \textbf{Base-Metric Initialization:} For experiments utilizing the identity initialization, the raw weights of the spline were initialized such that the resulting curve approximated $y = x$. The first control point was locked to the grid minimum ($-15.0$), and subsequent step sizes were initialized to $\ln(\exp(h) - 1.0)$, where $h$ is the grid spacing. The choice of spline initialization, either identity-based or random, is determined through inner-fold validation and a priori hypotheses about how model depth interacts with the geometric prior of the metric. This approach avoids data leakage while keeping the selection process consistent.
\end{itemize}

\subsection{Datasets}

\textbf{HDM05\citep{muller2007mocap}:} A 117-class kinematic action recognition dataset ($93 \times 93$ SPD matrices). Evaluated using 5-Fold Stratified Cross-Validation.\\
\textbf{FPHA\citep{garcia2018first}:} First-Person Hand Action dataset, containing irregular kinematic tracking data ($63 \times 63$ SPD matrices). Evaluated using the standard predefined train-test split.\\
\textbf{Radar\citep{brooks2019riemannian}:} A signal processing covariance dataset ($20 \times 20$ SPD matrices). Evaluated using 5-Fold Stratified CV.


\subsection{ Architecture configurations}
In all configurations, the output of the metric mapping layer was vectorized, passed through a \texttt{BatchNorm1d} layer to stabilize Euclidean convergence, and fed into a terminal Linear classifier. 

\textbf{1. Direct Linear Probe:} In order to gauge the intrinsic learning capability of a metric, it is imperative to isolate it from the compensatory effect of highly parameterized deep learning layers and classifiers. Hence, in this experiment, we project the uncompressed SPD matrices directly into the tangent space using respective metric before feeding them into a single linear layer. High accuracy here proves the metric itself has separated the classes.

\begin{itemize}
    \item \textbf{Flow:} $\text{Input} \to \text{Metric Projection} \to \text{Flatten} \to \text{BatchNorm1d} \to \text{Linear Classifier}$
    \item \textbf{Global AIRM Baseline:} Because AIRM lacks a standard direct projection, the Linear Probe for the AIRM baseline whitened the data using the Global Fréchet Mean of the training fold, $M_{global}^{-1/2} X M_{global}^{-1/2}$, before projecting to the tangent space via the matrix logarithm.
\end{itemize}

\textbf{2. SPDNet (SPD-MLR):}
We adopt the standard SPDNet building blocks \citep{huang2017riemannian}: Bilinear Mapping (BiMap) and Riemannian Eigenvalue Rectification (ReEig). Following recent advancements in generalized Riemannian classifiers \citep{chen2024rmlr}, the final classification layer of this backbone is implemented as an SPD Multinomial Logistic Regression (SPD MLR) to directly evaluate the performance of each metric mapping. To isolate the effect of spatial filtering from information bottlenecking, this architecture preserves the full-rank manifold topology without dimensionality reduction. We denote the architecture geometries by an array $[d_{in}, d_{out}]$.
\begin{itemize}
    \item \textbf{HDM05:} $[93, 93]$
    \item \textbf{FPHA:} $[63, 63]$
    \item \textbf{Radar:} $[20, 20]$
    \item \textbf{Architecture:} $\text{Input} \to \text{BiMap} \to \text{ReEig} \to \text{Metric Projection} \to \text{Flatten} \to \text{BatchNorm1d} \to \text{Linear Classifier}$
\end{itemize}

\textbf{3 Deep RResNet}
Following \citet{katsman2024riemannian}, the Riemannian Residual Network integrates residual connections within the manifold constraints. To test the metrics in bottlenecked environments, this architecture applies a dimensionality-reducing BiMap prior to the residual blocks.
\begin{itemize}
    \item \textbf{HDM05:} $[93, 30]$ 
    \item \textbf{FPHA:} $[63, 33]$ 
    \item \textbf{Radar:} $[20, 8]$ 
    \item \textbf{Architecture:} $\text{Input} \to \text{BiMap} \to \text{ReEig} \to \text{RResNet Block} \times 2 \to \text{Metric Projection} \to \text{Flatten} \to \text{BatchNorm1d} \to \text{Linear Classifier}$
    \item \textbf{AIRM Baseline Exception:} For the Deep RResNet configuration, the AIRM baseline utilized native AIRM RResNet blocks followed by a standard Log-Euclidean projection at the classification head, adhering to standard Riemannian ResNet practices, mirroring the official implementation.
\end{itemize}

Moreover, the $\theta$-PCM baseline \citep{chen2026fast} applies a fractional power to the diagonals of the Cholesky factor, scaled by an isometry constant. To grant the baseline maximum competitive advantage, we adopted the optimal $\theta$ values reported by Chen et al. for each dataset. 

However, because the original implementation relies on heavily dimension-reduced manifolds (like $[93 \to 30]$), applying these optimal $\theta$ values to the uncompressed, full-rank Linear Probe and SPDNet architectures frequently resulted in numerical divergence (NaNs) or \texttt{LinAlgError} exceptions due to the evaluation of fractional powers on microscopic eigenvalues. Our selection protocol was as follows: evaluate the author's optimal $\theta$; if the forward pass collapsed, iteratively evaluate the next-best reported $\theta$ until stability was achieved.
\begin{itemize}
    \item \textbf{Radar:} Linear Probe \& SPDNet ($\theta=0.5$), RResNet ($\theta=-1.0$).
    \item \textbf{HDM05:} All architectures ($\theta=0.5$).
    \item \textbf{FPHA:} Linear Probe \& SPDNet ($\theta=0.75$), RResNet ($\theta=0.5$).
\end{itemize}

\section{Theoretical vulnerabilities of rank-dependent matrix mappings}
\label{sec:alem_critique}

Recent works, such as the Adaptive Log-Euclidean Metrics (ALEM) \cite{chen2024adaptive}, propose learning parameterized metrics by applying distinct scalar functions to eigenvalues based on their sorted indices. For an SPD matrix $S = U \Sigma U^\top$, ALEM defines a rank-dependent matrix logarithm as:
\begin{equation}
    \phi_{ALEM}(S) = U \text{diag}(\log_{a_1}(\sigma_1), \dots, \log_{a_n}(\sigma_n)) U^\top
\end{equation}
where $a_i$ are independently learned base parameters for the $i$-th sorted eigenvalue. We demonstrate mathematically that this mapping is ill-defined (one-to-many) at spectral singularities and breaks standard backpropagation paradigms.

\subsection{Proof of ill-definedness via eigenspace non-uniqueness}
Consider a simple $2 \times 2$ SPD matrix $S$ with a repeated eigenvalue (a spectral singularity):
\begin{equation}
    S = \begin{bmatrix} 2 & 0 \\ 0 & 2 \end{bmatrix} = 2I
\end{equation}
Because $\sigma_1 = \sigma_2 = 2$, the eigenspace spans all of $\mathbb{R}^2$. The eigenvector matrix $U$ is not unique; any $2 \times 2$ orthogonal matrix is a valid basis. Consider two valid, solver-generated eigendecompositions $S = U_1 \Sigma U_1^\top$ and $S = U_2 \Sigma U_2^\top$, where:
\begin{equation}
    U_1 = \begin{bmatrix} 1 & 0 \\ 0 & 1 \end{bmatrix}, \quad U_2 = \frac{1}{\sqrt{2}}\begin{bmatrix} 1 & -1 \\ 1 & 1 \end{bmatrix}
\end{equation}
Both $U_1$ and $U_2$ perfectly reconstruct $S$. Now, assume the ALEM network has learned distinct parameters for the first and second eigenvalue ranks, such that $\log_{a_1}(2) = 1$ and $\log_{a_2}(2) = 2$. The transformed eigenvalue matrix is thus:
\begin{equation}
    \bar{\Sigma} = \begin{bmatrix} 1 & 0 \\ 0 & 2 \end{bmatrix}
\end{equation}
Reconstructing the output matrix using the first basis $U_1$ yields:
\begin{equation}
    Y_1 = U_1 \bar{\Sigma} U_1^\top = \begin{bmatrix} 1 & 0 \\ 0 & 2 \end{bmatrix}
\end{equation}
However, reconstructing the output matrix using the second valid basis $U_2$ yields:
\begin{equation}
    Y_2 = U_2 \bar{\Sigma} U_2^\top = \frac{1}{2} \begin{bmatrix} 1 & -1 \\ 1 & 1 \end{bmatrix} \begin{bmatrix} 1 & 0 \\ 0 & 2 \end{bmatrix} \begin{bmatrix} 1 & 1 \\ -1 & 1 \end{bmatrix} = \begin{bmatrix} 1.5 & -0.5 \\ -0.5 & 1.5 \end{bmatrix}
\end{equation}
We observe that $Y_1 \neq Y_2$. The same input matrix $S$ has been mapped to two entirely different output matrices. Because the transformation breaks the isotropic symmetry of the repeating eigenvalues, the arbitrary rotation chosen by the computational solver does not cancel. Therefore, ALEM is a one-to-many mapping, failing the fundamental definition of a mathematical function.

In contrast to rank-dependent approaches, our proposed method (SPM) avoids these failures by strictly adhering to the calculus of Primary Matrix Functions. SPM applies a universally rank-invariant, strictly monotonic scalar spline $f_\theta: \mathbb{R}_{>0} \to \mathbb{R}$ to all eigenvalues equally, independent of their sorted index.

\subsubsection{Well-definedness at spectral singularities}
Revisiting the problematic $S = 2I$ case with our proposed SPM, the eigenvalues are identically $\sigma_1 = \sigma_2 = 2$. Because our function is rank-invariant, it evaluates to a single scalar $c = f_\theta(2)$ for both dimensions. The transformed eigenvalue matrix becomes a scalar multiple of the identity:
\begin{equation}
    \bar{\Sigma}_{SPM} = \begin{bmatrix} c & 0 \\ 0 & c \end{bmatrix} = cI
\end{equation}
When we reconstruct the output matrix using the two  different, solver-generated eigenvector bases $U_1$ and $U_2$, we obtain:
\begin{align}
    Y_1 &= U_1 (cI) U_1^\top = c (U_1 U_1^\top) = cI \\
    Y_2 &= U_2 (cI) U_2^\top = c (U_2 U_2^\top) = cI
\end{align}
Because $U U^\top = I$ is a universal axiom for any orthogonal matrix, the arbitrary rotations mathematically cancel themselves. We achieve $Y_1 = Y_2 = cI$.

\subsection{Topological constraints of Cholesky Power Metric (PCM)}
Recent methods have attempted to avert the computational cost of spectral decompositions by utilizing the Cholesky manifold, notably the Power-Cholesky Metric (PCM) \cite{chen2026fast}. PCM defines a product geometry where the strictly lower triangular elements of the Cholesky factor $L$ remain Euclidean, while a power function $x^\theta$ is applied to the positive diagonal elements.

While this avoids eigenvector ambiguity, it introduces a severe topological limitation: a lack of global surjectivity onto the unconstrained tangent space.

To allow for unconstrained network optimization, a valid Riemannian mapping must project the SPD manifold diffeomorphically onto the entirety of the unconstrained Euclidean space $\mathbb{R}$. The matrix logarithm achieves this because the scalar logarithm surjectively maps $(0, \infty) \to (-\infty, \infty)$. 

In contrast, PCM relies on the mapping $\phi_\theta(L_{ii}) = L_{ii}^\theta$. For any $L_{ii} > 0$ and $\theta > 0$, the image is strictly $(0, \infty)$. Consequently, the Riemannian logarithm for PCM on the diagonal is structurally bounded. As defined by the authors, for diagonal elements $K_{ii}, L_{ii} > 0$:
\begin{equation}
    \text{Log}_L(K)_{ii} = \frac{1}{\theta}L_{ii} \left[ \left(\frac{K_{ii}}{L_{ii}}\right)^\theta - 1 \right]
\end{equation}
As $K_{ii} \to 0$, the tangent vector is strictly bounded from below by $-\frac{1}{\theta}L_{ii}$. It is mathematically impossible for the PCM logarithm to map to tangent vectors that are more negative than this bound. Because the mapping is not surjective onto $\mathbb{R}$, PCM does not constitute a true global diffeomorphism to an unconstrained space, forcing its geodesics to be only locally defined.

Because the tangent space is bounded, unconstrained updates may project the diagonal elements outside the valid manifold domain. To prevent computational failure, the authors propose a ``numeric trick'' (Remark 3.9 in \cite{chen2026fast}) utilizing a hard clamp: $d_i \leftarrow \max(d_i, \epsilon)$.

This clamping operation breaks the geometric validity of the metric. The $\max$ function is a non-injective (many-to-one) operator; any pre-image value $x \le \epsilon$ is mapped to the same output. Thus, bijectivity is lost. The gradient of this clamping operation evaluates to zero for all $x < \epsilon$, limiting backpropagation. 

\subsubsection{Resolution via surjective spline generators (SPM)}
Our proposed method avoids these topological failures.  SPM, being a vaild diffeomorphism, achieves an unconstrained tangent space, inducing geodesic completeness. The network can output any arbitrary real number in the tangent space without requiring non-injective clamping operations establishing a mathematically sound framework for SPD matrix learning.

\section{Limitations and Future Directions}
\label{sec:appendix_limitations}

While the Spline-Pullback Metric (SPM) eliminates the limitations of rigid mathematical priors, its high expressivity introduces specific limitations that invite future research. First, the expressivity of the diffeomorphism relies on the predefined B-spline grid resolution, spline order, and spectral boundaries. We mitigate this sensitivity in our current framework by fixing the grid bounds for all the experiments without using the dataset for further grid search or tuning to avoid optimization bias, which is glaringly prevalent in the literature. Furthermore, SPM's bounded linear extrapolation acts as a safety net to process any out-of-distribution spectral singularities without requiring exhaustive hyperparameter grid searches. Future work, however, will explore automated, data-driven knot placement, like utilizing the eigenspectrum information of the train and validation sets of the specific fold. Additionally, non-linear metric tensors may be sensitive to their initial geometric state. We currently resolve this concern of early-epoch instability by introducing specific Base-Metric (Identity) and Random initializations for stable convergence. Future research in Riemannian Metric Initialization (RMI), as stated before in the Discussion and Conclusion section, will expand upon this by deriving initialization distributions tailored specifically to the SPM manifold. Also, the computational learning requirement of the SPM, as opposed to predefined metrics, might appear as a computational limitation; however, this concern is more than offset by the time and computational need to manually choose and test the optimal metric for the specific dataset, which SPM learns automatically. The single-layer, parameter-efficient C-SPM further mitigates computational concerns by averting expensive eigendecompositions.

Secondly, as a universal diffeomorphism approximator, SPM possesses the theoretical capacity to overfit in data-scarce domains. While our empirical evaluations did not suffer from any pronounced overfitting, this limitation motivates the development of Riemannian Metric Regularization (RMR). Rather than relying solely on standard Euclidean weight decay, future RMR techniques will aim to explicitly ensure the integrity of the topological and diffeomorphic rules of the manifold. 
Finally, since anonymized datasets, provided by the original authors, were utilized in this work, no privacy or ethical issues arise.

\newpage
\end{document}

%% file: math_commands.tex
\usepackage{amsmath,amsfonts,bm}

\def\eqref#1{equation~\ref{#1}}

\def\1{\bm{1}}

\DeclareMathAlphabet{\mathsfit}{\encodingdefault}{\sfdefault}{m}{sl}
\SetMathAlphabet{\mathsfit}{bold}{\encodingdefault}{\sfdefault}{bx}{n}